\documentclass[sigconf]{acmart}

\usepackage{algorithmic}
\usepackage{graphicx}
\usepackage{textcomp}
\usepackage{filecontents}
\usepackage{multirow}
\usepackage{makecell}
\usepackage{xcolor}
\usepackage{etoolbox}
\usepackage{hyperref}
\makeatletter

\renewcommand{\ceur}{EEKE 2021 - Workshop on Extraction and Evaluation of Knowledge Entities from Scientific Documents}
\renewcommand{\ceurLicense}{Copyright 2021 for this paper by its authors. Use permitted under Creative Commons License Attribution 4.0 International (CC BY 4.0).}

\AtBeginDocument{%
  \providecommand\BibTeX{{%
    \normalfont B\kern-0.5em{\scshape i\kern-0.25em b}\kern-0.8em\TeX}}
}

\begin{document}

\title{ANEA: Automated (Named) Entity Annotation for German Domain-Specific Texts}

\author{Anastasia Zhukova}
\email{zhukova@uni-wuppertal.de}
\orcid{0000-0001-9084-2890}
\authornotemark[1]
\affiliation{%
  \institution{University of Wuppertal}
%   \city{Wuppertal}
  \country{Germany}
}

\author{Felix Hamborg}
\email{felix.hamborg@uni-konstanz.de}
\authornotemark[2]
\affiliation{%
  \institution{University of Konstanz}
%   \city{Konstanz}
  \country{Germany}
}

\author{Bela Gipp}
\email{gipp@uni-wuppertal.de}
\authornotemark[1]
\affiliation{%
  \institution{University of Wuppertal}
%   \city{Wuppertal}
  \country{Germany}
}

% \renewcommand{\shortauthors}{Zhukova et al.}

\begin{abstract}
Named entity recognition (NER) is an important task that aims to resolve universal categories of named entities, e.g., persons, locations, organizations, and times. Despite its common and viable use in many use cases, NER is barely applicable in domains where general categories are suboptimal, such as engineering or medicine. To facilitate NER of domain-specific types, we propose ANEA, an automated (named) entity annotator to assist human annotators in creating domain-specific NER corpora for German text collections when given a set of domain-specific texts. In our evaluation, we find that ANEA automatically identifies terms that best represent the texts' content, identifies groups of coherent terms, and extracts and assigns descriptive labels to these groups, i.e., annotates text datasets into the domain (named) entities. 
\end{abstract}


% TODO 
\begin{CCSXML}
<ccs2012>
<concept>
<concept_id>10002951.10003317.10003347.10003352</concept_id>
<concept_desc>Information systems~Information extraction</concept_desc>
<concept_significance>300</concept_significance>
</concept>
<concept>
<concept_id>10010147.10010178.10010179.10003352</concept_id>
<concept_desc>Computing methodologies~Information extraction</concept_desc>
<concept_significance>300</concept_significance>
</concept>
<concept>
<concept_id>10010147.10010178.10010179.10010186</concept_id>
<concept_desc>Computing methodologies~Language resources</concept_desc>
<concept_significance>300</concept_significance>
</concept>
<concept>
<concept_id>10010147.10010257.10010258.10010260.10003697</concept_id>
<concept_desc>Computing methodologies~Cluster analysis</concept_desc>
<concept_significance>300</concept_significance>
</concept>
</ccs2012>
\end{CCSXML}

\ccsdesc[300]{Information systems~Information extraction}
\ccsdesc[300]{Computing methodologies~Information extraction}
\ccsdesc[300]{Computing methodologies~Language resources}
\ccsdesc[300]{Computing methodologies~Cluster analysis}
\keywords{information extraction, low-resource languages, named entity recognition, domain-specific texts}


\maketitle
%insert this line directly after maketitle
% \thispagestyle{firststyle}
% \newpage
% optional re-enable header on the second page (do not execute this command on the first page)
% \pagestyle{standardpagestyle}
% \maketitle

\section{Introduction}
Named entity recognition (NER), a common preprocessing step in natural language processing (NLP) for various tasks, such as information extraction, summarization, question and answering, and text understanding, is often criticized for capably representing only datasets with few general categories, e.g., person, location, organization, and time (including their subcategories) \cite{zhang2013unsupervised, eltyeb2014chemical}. While the original NER task contains only few categories, a rapidly increasing number of NER applications show a high demand for the datasets with \textit{domain-specific named entities} \cite{rocktaschel2012chemspot}. 

% Creation of the domain-specific datasets requires costly manual annotation. Training supervised NER systems requires often large datasets but their creation is a time-consuming process. 
To (semi-)automatically create large general-purpose NER corpora, recent research projects extensively use structured domain sources, such as dictionaries, knowledge graphs, and Wikipedia or other knowledge bases \cite{richman2008mining, simon2012automatically, zhou2015automatically}. 

In this paper, we propose ANEA, an unsupervised Wiktionary-based approach that automatically derives domain entities from German texts, i.e., low-resource language, by (1) extracting terms from topically-related domain texts, (2) identifying the most domain-representative, i.e., semantically distinct, terms of the analyzed texts, and (3) automatically annotating the terms, i.e., ANEA extracts labels from Wiktionary and assigns them to the identified groups of terms. Not all of the domain categories may be named, e.g., machinery or process.

By automating the most labor-intense parts, the proposed unsupervised approach minimizes the cost of expensive and laborious annotations required for the creation of domain-specific NER datasets. Typical manual tasks in annotations include (1) reading the domain text multiple times, (2) deriving entities based on the text content, and (3) manually selecting terms that match the derived categories. ANEA substitutes the most time-consuming task of deriving a coding book and automatically defines categories and annotates the most representative terms (nouns) into these categories. We evaluate the approach with user studies on multiple domain datasets against multiple silver datasets and discuss a default input configuration for ANEA to annotate other domain NER datasets\footnote{ \url{https://github.com/anastasia-zhukova/ANEA}}.

\section{Related work}
% no domain tagging, only annotation 
NER datasets usually contain standard types, e.g., person, location, organization, and are manually annotated \cite{sang2003introduction, benikova2014nosta} or automatically extracted \cite{richman2008mining, simon2012automatically, zhou2015automatically}. 
% , and \textit{resolution (NER)}, i.e., automatic identification of the NE spans in text and assignment of the category labels, e.g., for German language  \cite{rocktaschel2012chemspot, loster2017improving, ruppenhoferfine2020}.
Domain-specific NER typically needs to introduce domain-specific (sub-)categories of the established named entity (NE) categories or entirely new categories. This is because domain-specific texts contain NE categories that are  (1) detailed variants of the standard NE categories, e.g., ``Person'' is replaced with the domain-specific sub-categories ``Players'' and ``Coaches'' \cite{weber2006web}, (2) standard NE categories extended with a small number of new categories, e.g., ``Trigger of a traffic jam'' \cite{schiersch2018german, loster2018curex, ruppenhoferfine2020}, and (3) domain-derived NE categories, e.g., ``Proteins'' in biology or ``Reactions'' in chemistry domains \cite{rocktaschel2012chemspot, zhang2013unsupervised, heuss2014comparison, tomori2016domain}. Most domain-derived NE categories originate from structured classifications or dictionaries \cite{heuss2014comparison, tomori2016domain,  loster2017improving} or are derived by manually unifying multiple of them \cite{eltyeb2014chemical}. In sum, creating domain-specific datasets for NER requires expert knowledge and is time-consuming. 

To minimize such efforts, some NER approaches use seed-NEs, i.e., a small number of manually provided terms and their NE-categories \cite{nadeau2006unsupervised, zhang2013unsupervised, foley2018named}.
Such approaches use the seed-NEs as examples to extract patterns of NE definitions and apply them to the full text suggested for annotation. %The seed-based NER systems often rely on the domain knowledge bases (KB) (e.g., medical) or lexical bases, e.g., WordNet. 
% The typically underlying domain knowledge bases (KB), e.g., medical, ensure domain undertandng of such NER systems. On the contrary, such systems suffer from the slow-pace KB update and lead to low recall on the texts with unknown terms. 
These NER approaches suffer from the slow updates of the underlying domain knowledge bases (KB) \cite{yadav2018survey} and perform worse on lower-resource languages than on English \cite{heuss2014comparison}.
An alternative to domain-KBs is community KBs, such as Wikipedia and Wiktionary, which are constantly updated by their communities. They prove to contain a sufficient amount of domain information \cite{zesch2008extracting, meyer2011psycholinguists}.

% Wikipedia has also proven to be a valid resource to creation large silver, i.e., automatically created, datasets for NER of standard types (person, location, organization) in many languages \cite{richman2008mining, simon2012automatically, zhou2015automatically}.

Unlike the existing supervised approaches for annotating domain-specific named entities \cite{mansouri2008named, etzioni2005unsupervised}, in this paper, we explore ANEA, an unsupervised method to support researchers and users during the creation of a coding book. Given a set of domain- or use case-specific documents, ANEA automatically derives domain-specific categories and exemplary terms within. This way, ANEA automates the most time-intensive, previously manual tasks. As a consequence, users only need to revise these terms, e.g., by renaming the categories or re-annotating the not matching to the categories terms.
% To the best of our knowledge, only \citet{etzioni2005unsupervised} proposed and tested a syntactic-based system to extract terms and assign domain-NE-labels. 

\section{Methodology}
We propose an unsupervised approach for annotation of domain-specific (named) entities (ANEA) on a lower-resource language. The goal of ANEA is to fully automatically derive entity categories (later in the text: categories) by selecting groups of related terms and extract and assign a meaningful label to these terms. To do so, ANEA, first, links terms extracted from domain-specific texts to pages in Wiktionary \cite{meyer2011psycholinguists, simon2012automatically}. Second, ANEA automatically identifies groups of related terms and automatically labels them by performing a double optimization task of both maximizations of cross-similarity of terms in a group and the average similarity of these terms to a candidate label. That is, the approach consists of two main steps: (1) text preprocessing, i.e., term mapping to Wiktionary pages (WPs) and construction of a domain graph, and (2) identification of related terms and label assignment.

\subsection{Preprocessing and domain graph}
\label{sec:prepr}
\subsubsection{Preprocessing}  
\begin{figure}[h]
\centering
\includegraphics[width=0.45\textwidth]{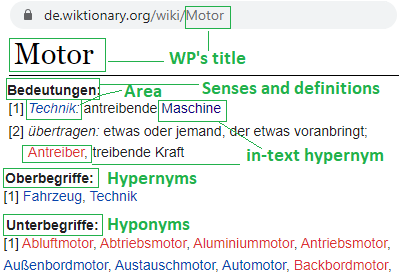}
\caption{\label{img:example} An example of a Wiktionary page (WP).}
\end{figure}

The goal of preprocessing is to extract terms from the set of texts and maximize the number of terms aligned to the Wiktionary structure, i.e., map the domain-specific terminology to the structured knowledge base. The mapping of the extracted terms to the knowledge graph enables using their semantic information, such as term definitions, areas, hypernyms, and hyponyms (see Figure \ref{img:example}).

The preprocessing steps include parsing and part-of-speech (POS) tagging using spaCy \cite{spacy2}.  We define a \textit{term} as any unique noun phrase that does not contain any digits \cite{foley2018named}. After extraction, terms are mapped to their respective German WP, if any, i.e., each term gets assigned to a link of a WP. 

In German texts, we find that many domain-specific terms are compound words, i.e., words that consist of more than one noun component, for example, ``Sechszylindermotor'' = ``sechs'' + ``Zylinder'' + ``Motor'' (six-cylinder motor).
% (F: nice example :-))
Typically, such complex domain compound words are not described in Wiktionary since they are too rare or specific. On the contrary, we observe that compound words' \textit{heads}, i.e., the part of a composition term that bears the core meaning of the phrase, e.g., ``Motor'' in the example above, are highly likely to have a WP in Wiktionary. 

\begin{figure*}[h]
\centering
\includegraphics[width=0.95\textwidth]{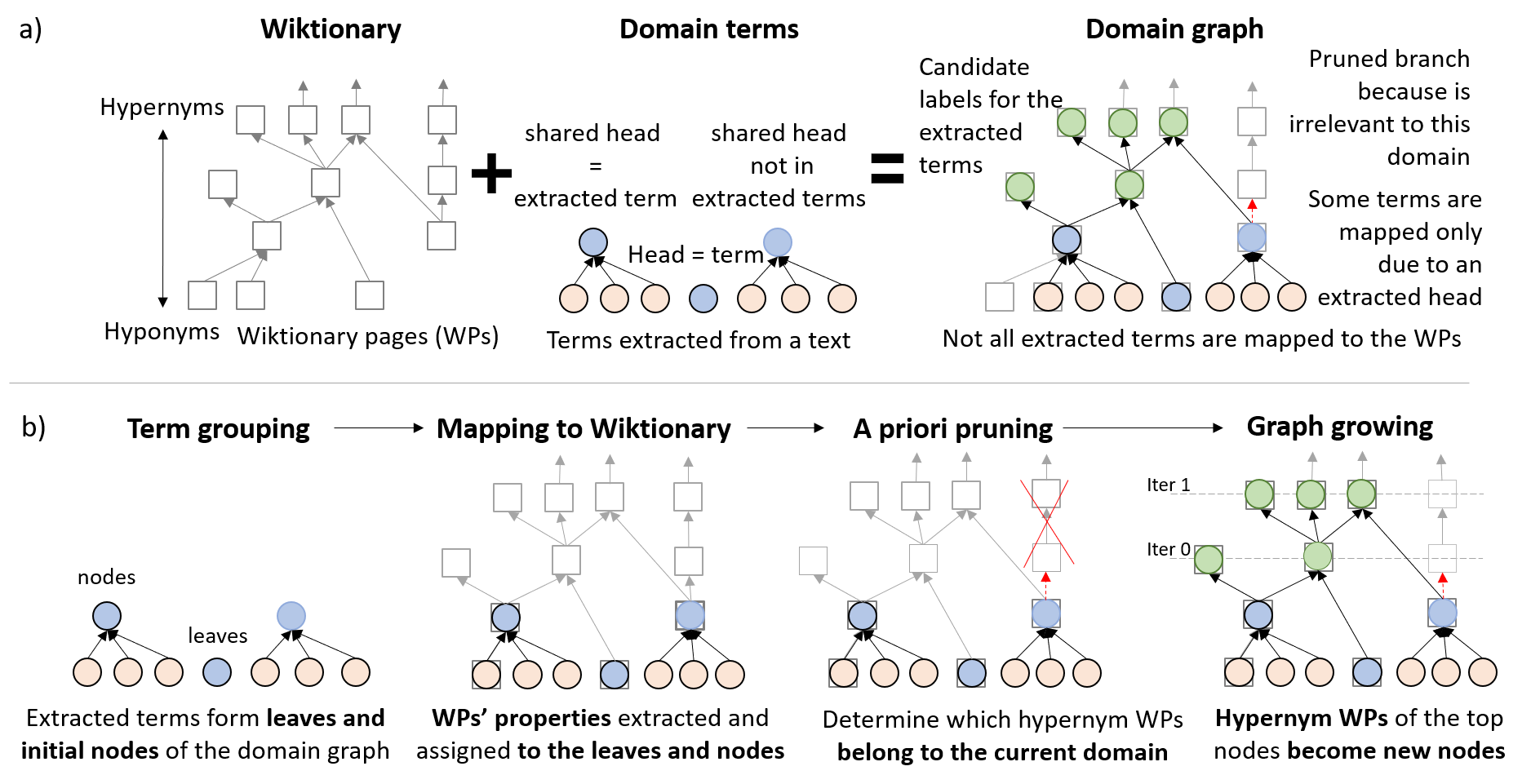}
\caption{\label{img:wikt} (a) A domain graph is a combination of Wiktionary and domain terminology extracted from text. 
(b) Domain graph construction: initialization and expansion of the graph.}
\end{figure*}

To map rare domain-specific terms to WPs, we extract the heads of the extracted terms with a compound splitter, i.e., a model that splits terms into two parts, compound- and head-parts, \cite{Tuggener2016} and attempt to map the heads to Wiktionary. If a (multi-token) term has a corresponding WP, we set a full term as a term's head.  If the compound splitter outputs a head that is not a part of Wiktionary, we continue recursively search for a head that can be mapped to a WP. If no heads have corresponding WPs, then we do not assign a head to a term. 

% TODOA
The preprocessing could be changed to include the terms with digits, but for now, we focused on the noun phrases as terms. If a term or its head of compound phrase do not have a WP, they are excluded from annotation because the absence of a link to Wiktionary leads to an inability to map terms to potential category labels. Later, such discarded terms can be manually classified by human annotators or automatically with state-of-the-art NER models trained on the automatically created domain datasets. 
 
We use fastText to vectorize extracted terms and candidate labels \cite{grave2018learning}. We chose fastText due to its ability to vectorize out-of-vocabulary words, which often happens with domain-specific terminology. 

% \subsection{Domain graph}

\subsubsection{Domain graph} The \textit{domain graph} is a locally stored knowledge graph where the leaves are the extracted domain terms. Nodes are all terms obtained from the WPs linked to the leaves and to each other with hyponymy-hypernymy relations. Figure~\ref{img:wikt}a depicts the principle of a domain graph. The construction of the domain graph includes three steps: (1) graph initialization, i.e., extraction of the WP properties, e.g., definitions, with the scraping of the WPs assigned to the domain terms and their heads; (2) determination of pruning criteria of Wiktionary graph to scrape only domain related pages; (3) expansion of the domain graph, i.e., scrapping of the hypernym pages, to create a pool of candidate labels to later annotated the identified groups of terms. Figure~\ref{img:wikt}b shows the process of domain graph construction.

\subsubsection{Initialization}
% (F: add 1-2 sentences: what is the input to the initialization step, what will be the output, and why is it necessary; similar as you did for the a priori pruning :-) )
To initialize the graph, we use the extracted terms to which we mapped WPs and scrap the mapped WPs to extract  WPs' properties, e.g., hypernyms. As a preliminary step of the graph initialization, we group the extracted terms by their head. The \textit{head grouping} aims at the extraction of the initial hyponym-hypernym relation for the domain graph. Then, we sort the list of heads in decreasing order by (1) the number of unique terms with each head, (2) the frequency of the overall in-text occurrence of words with such head. 

To maximize the descriptiveness and generalization of the terms that will become annotated into categories, we initialize the domain graph with the terms-to-annotate that belong to the top $M$ largest head groups, i.e., containing the most lexically diverse and/or frequent terms. Section~\ref{sec:eval} determines an optimal value of terms-to-annotate and the largest head groups the series of experiments. This filtering procedure reduces the size of the domain graph to minimize the time of the execution and extract the most representative candidate labels, i.e., the most closely located hypernyms. 
% As opposed to selecting the most frequent extracted terms to initialize a graph, the selection of the frequent groups of heads guarantees that most of representative terms  due the high frequency  and initial relations between terms will become a part of the domain graph. 

Each term without a hyponym is a \textit{leaf} of the domain graph; a \textit{node} is a head that aggregates more than one term. We scrape WPs of all leave- and node-terms to extract the text and links from definitions, hypernyms and hyponyms (see Figure~\ref{img:example}). 
% (F: just fyi, you can use the cleverref package to use backslashCref to not need to write Figure or Table :-))

We extract hyponym terms from the corresponding WP's section.
% (F: i suggest to use the english terms here. i would suggest to keep the description conceptual, because in principle this approach would work for any language that has a lot of compounds; so i would abstract this from technical (what exactly is the name of the section on the webpage) to conceptual (which section is the one we are interested in semantically))
We extract hypernym terms from two WP's parts: (1) the hypernym section,  (2) the definition section by parsing the text of term's definitions and ensuring that the extracted word has its WP\footnote{We extract the tokens that have one of the following dependency tags: ``ROOT'', ``oa''=	accusative object, ``oa2'' = second accusative object, ``app'' = apposition, ``cj'' = conjunct.}. For example, in Figure~\ref{img:example}, the word ``Maschine'' will be extracted as an additional (in-text) hypernym to those listed in the hypernym section.

The extracted properties are assigned to each node. The hypernyms' links point at the WPs that may later become nodes of the domain graph. 
% we do not instantly scrap the WP of the extracted hypernyms and hyponyms and add them as nodes. 
Extraction and assignment of the WP's properties bridge the domain terms and heads to the Wiktionary's knowledge graph.

\subsubsection{A priori pruning}
Most of the terms of WPs have more than one sense and some of them may be associated with the different semantic areas, e.g., technology, medicine, sport, law, etc. If a sense belongs to only one area, the title of the area precedes the definition explaining the sense, e.g., ``Technik'' in Figure~\ref{img:example}. The step \textit{a priori pruning} determines which senses of yet to add hypernyms need to comply with the senses of the previously added terms.
% (F: i understand what this does, but i think the phrase "determines the selection criteria" is not optimal. maybe better remove it eg: The step \textit{a priori pruning} determines which senses of hypernyms comply with the senses of the previously added terms)
% for hypernyms in the domain graph, i.e., which senses of hypernyms comply with the senses of the previously added terms. (F: not sure about the following sentence, but it was not grammatically correct, tried to rewrite it but unsure whether it still has the meaning you intended)
Hypernyms will become properties of a node in the domain graph if and only if their areas belong to a predefined list of areas or if they do not list any domain areas.

To identify which Wiktionary's areas determine the graph's domain, we select the most frequent and semantically similar areas extracted from the senses' definitions of the previously added leaves and nodes. To find the most semantically similar and frequent areas, we cluster all titles using hierarchical clustering and select the areas from the most representative clusters. 
% (F: i think you really like clustering :-) it is in every paper we work on :-D)
As parameters of the hierarchical clustering, we use Euclidean distance, average linkage criterion, and optimize the number of clusters. To represent areas' titles in the vector space, we apply the fastText word embeddings model \cite{grave2018learning}.

We extract three most representative clusters by: (1) selecting a cluster with the average cosine similarity across all words in a cluster being the highest among all clusters ($C_s$); (2) selecting a cluster with the count of all words in a cluster is the highest among all clusters ($C_f$); (3) forming an extra cluster with the $K$ most frequent areas  $C_k$ ($K = 5$). To identify the Wiktionary areas $A$ forming the domain of the graph, we intersect all three representative clusters: $ A =  \{\cup a: a \in C_s \cap C_f \cap C_k\} $. Finally, we select a clustering configuration that outputs the best domain-defining areas as $A_{best} = \operatorname{arg\,max}_{6\leq i \leq 12} (s(A_i) \cdot f(A_i))$, where $A_i$ are the areas identifies at $i$ the number of clusters, $s(A_i)$ is an average cross similarity of the areas in $A_i$ and $f(A_i)$ is sum of $A_i$' frequencies.

\subsubsection{Graph growing}
The goal of NEA is to assign the most generalizing yet still representative labels to groups of semantically related terms, e.g., ``Person'' for ``Trump'' and ``Einstein.'' To ensure the generalization property of label candidates, we \textit{``grow''}, i.e., expand the domain graph up, by adding new nodes on the top of the graph from the scrapped hypernym WPs. 

To grow the graph, we iterate over the top nodes and create new nodes for each of the hypernym terms. To obtain node's properties, we scrape WPs of the hypernym terms and extract term definitions, hyponyms, and hypernyms. For each new node, we add hypernym-hyponym edges between this new node and the matching previously added nodes while also removing any edges creating cycles in the domain graph. To avoid over-generalizing candidate labels, we perform only one or two iterations of the graph growing.

\subsection{Automated term grouping and labeling}
\label{sec:anea}
The goal of ANEA is to obtain (named) \textit{entity categories}, i.e., few clusters of generally related terms of high cross-similarity and assign descriptive labels to these clusters. To do so, we maximize two parameters at the same time: cross-term group similarity and similarity between a group of terms and a label.

ANEA consists of the initial setup of the categories and three subsequent optimization steps to improve the representativeness of the terms and the assigned labels in the groups of terms.

\subsubsection{Setup} 
\begin{figure}[h]
\centering
\includegraphics[width=0.4\textwidth]{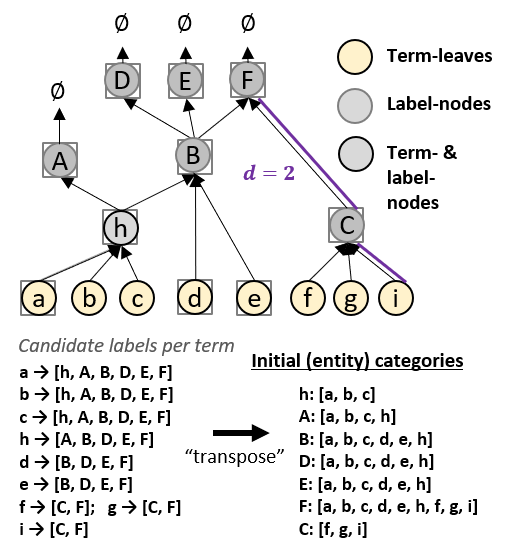}
\caption{\label{img:anea}  ANEA setup: a collection of candidate entity categories. }
\end{figure}

We initialize ANEA by collecting all candidate categories, i.e., groups of potentially related terms and the assigned labels to them. Figure~\ref{img:anea} depicts the process of candidate label collection.

\begin{figure}[h]
\centering
\includegraphics[width=0.45\textwidth]{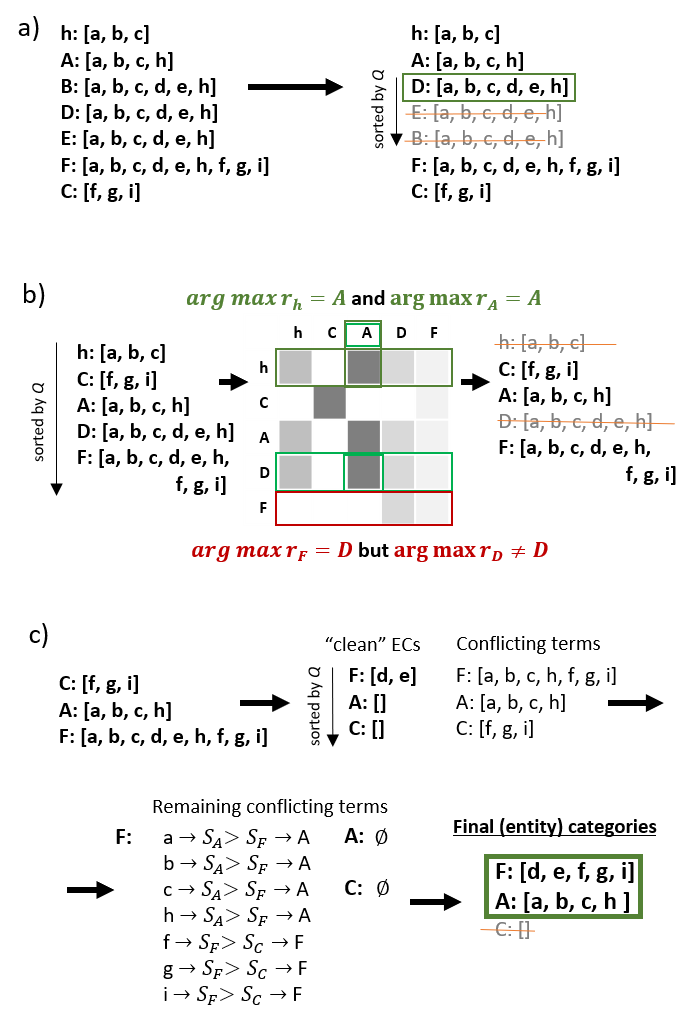}
\caption{\label{img:steps} \centering (a) Resolution of the full overlaps. 
 (b) Resolution of the substantial overlaps. (c) Resolution of the conflicting terms. }
\end{figure}

First, we iterate over all term-nodes, i.e., the domain graph's leaves and nodes that were created from the extracted terms not the hypernym WPs. For each term-node, we collect candidate labels extracted from the names of their hypernym-nodes. Each term obtains a list of candidate labels with various \textit{distances}, i.e., the number of edges between a term-node and a label-node. We recursively traverse the domain graph as long as the distance between a term-node and a label-node is $d \leq d_{max}$, $d_{max}=5$, or there are hypernyms to the current node in the domain graph. During the experiments, we notice that label-nodes with larger distances are often rather abstract and do not characterize a term well.  

Second, we ``transpose'' all terms and their candidate labels to obtain one label assigned to a group of terms. That is, we create a collection of categories among which we seek to find the most representative categories of the analyzed domain-specific text collection.

The selection of the most optimal categories among the candidates is a double-optimization process towards two requirements: generalization and specification.  On the one hand, \textit{generalization} aims at covering categories' broader semantics, e.g., a category with a more general label ``Person'' is better than categories such as ``Actor,'' ``Politician,'' etc. On the other hand, \textit{specification} aims at selecting the category with more narrow semantics, e.g., categories such as ``Country,'' ``City,'' ``State'' provide more details about its terms than a category ``Location''. 

We use a \textit{quality score} $Q_i$ to evaluate each (entity) category $EC_i$ in a list of candidates: % F. i like the equations! :-)
\[ Q_i = T_i \cdot L_i \cdot  O_i \cdot \operatorname*{max}({\log_{2} |EC_i|, 1 }) \cdot d_{avg\_i} \]
where $T_i$ is a mean cross-term cosine similarity; $L_i$ is a mean label-terms similarity; $O_i$ is the overall similarity, i.e., $O_i = T_i + L_i$; $|EC_i|$ is a size of a category, i.e., the number of the terms in the class;  $d_{avg\_i}$ is an average of non-zero distances between category's terms and label $l$: $d_{avg\_i} = \frac{1}{|D_{nn\_i|}} \sum_{d \in D_{nn\_i}} d $, where $D_{nn\_i} = \{D_{i,l}| \, \forall i \in EC: D_{i,l} > 0\}$ and $D$ is a distance matrix\footnote{Rows are the names of term-nodes and the columns are the names of term-nodes and label-nodes. The columns contain also the term-nodes because some of the term-nodes may not be leaves but nodes of the domain graph (see Figure~\ref{img:wikt}). }. If $|D_{nn\_i}|= 0$, then $d_{avg\_i}=1$.

To calculate cosine similarities, we represent each term and label in a vector space with fastText word embeddings \cite{grave2018learning}. We chose fastText for the representation of the out-of-vocabulary words, which often occur in domain-specific texts.
% \footnote{We plan to use ELMo and other context-depended word embeddings in future work. In this work, we experimented with the static semantic representation of each term.  }.
% this footnote could be removed, esp in case we need space, because it kind of is a trivial statement and doesnt really add value to the paper

Requiring a large mean cross-term cosine similarity $T_i$ increases the specificity of a category. Typically, the smaller the number of related terms, the larger the mean cross-similarity is. A larger mean label-terms similarity $L_i$ also increases the specificity, i.e., a large similarity value is equivalent to a narrow descriptiveness of the terms of a label. 

The overall sum $O_i$ facilitates balancing potentially small values of either $T_i$ or $L_i$ if the other item is still large. The large size of a category increases its generalizing and descriptive properties, i.e., one label needs to describe as many terms as possible. Lastly, the average distance $d_{avg}\_i$ acts as an amplifying factor for the generalization: the higher the label $l$ is in the domain graph, the more general is its meaning to the terms in this category.

Before the optimization steps, we perform filtering of the candidate categories to remove low quality categories from the candidate list. We remove an $EC_i$ if: (1) $T_i < 0.2$, (2) $L_i < 0.3$, (3) $|EC_i| > 0.15 \cdot |TTA| \land |EC_i| < 5$, where $|TTA|$ is a number of terms-to-annotate, i.e., a number of the term-nodes in the domain graph. In other words, we remove too vaguely related, very large or small categories.

\subsubsection{Resolution of full overlaps} 
Figure~\ref{img:steps}a depicts that if two categories have the same terms but different labels, we sort the classes by their quality scores $Q_i$ and keep the categories with the highest $Q_i$.

\subsubsection{Resolution of substantial overlaps}
%F: rather than describing some characteristics i would recommend to describe a problem that motivates why we have this optimization step that is now being described.
Typically, categories have overlaps between their terms albeit we find that cross-terms and terms-label combinations of a single category are more semantically coherent than combinations of another category. We define that categories have substantial overlap if they share more than 50\% of their terms. 

Figure~\ref{img:steps}b depicts the process of conflict resolution. We construct a matrix of replacements $R$, i.e., a matrix indicating the quality of a category measured by $Q$ compared to those categories with substantial overlaps (values of $R$ initialized with 0). The matrix is used to identify if an $EC_A$ contains the best terms-label combination or there is a better $EC_{repl}$ to replace $EC_A$. Since ANEA's goal is to annotate as many generally related terms as possible yet find as specific categories as possible, we challenge both the size of $EC_a$ and its descriptive properties.

First, we sort categories by their $Q$ score in decreasing order. 
Second, we intersect all categories with each other.
% Second, we iterate in a double loop over the entity classes and overlap the terms from the outer loop $A$ (which also denotes all entity classes) with the inner loop $B$. 
If $|EC_a \cap EC_b| \geq 0.5 \cdot |EC_a|$, then we consider the overlaps substantial and add the quality score $Q_b$ to the matrix of replacements as a $R_{a,b}$ value. Note that the matrix is squared but asymmetrical because we calculate 50\% of $EC_a$'s size and not of a pairwise function of two categories, e.g., $\operatorname*{min}(|EC_a|, |EC_b|)$. 
%F: uff, this might sound a bit to technical / close to the actual implementation (eg "double loop", i think it be good to remove these programming terms and try to abstract <way from the implemnentation and conceptually describe it :)

Finally, for each $EC_a: \forall a \in A$ represented by a row $r_a$ in $R$, we select a replacement $EC_{repl}$: 
\begin{multline*}
EC_{repl}(EC_a) = \{EC_c | \quad \exists c \in A :\\  \quad \operatorname*{arg\,max} r_a = EC_c\, \land     \operatorname*{arg\,max} r_c = EC_c \}
\end{multline*}
That is, we call $EC_c$ a replacement to $EC_a$ if $EC_c$ is the best among all comparable categories to $EC_a$ and also $EC_c$ is the best among all categories compared to itself. Also, a category can be a replacement for itself. We keep only the unique categories that are the best replacements $\{EC_{repl}(E_a): \forall a \in A\}$.

\subsubsection{Resolution of conflicting terms} 
After resolution of substantially overlapping terms, some categories contain minor conflicting terms, i.e., that are present in more than one category (Figure~\ref{img:steps}c). 
% Figure~\ref{img:anea}d depicts the last optimization step ensures a one-to-one assignment of the terms already used in EC to the categories. 

To resolve conflicting terms, first, we create a list of ``clean'' categories, i.e., from each category we remove all conflicting terms and record categories' labels from which the conflicting terms were removed. Additionally, we resolve the terms of the categories that may be also labels of another category, e.g., such a  term as ``h'' in Figure~\ref{img:anea}, i.e., move these label-terms to the categories with corresponding labels. We keep all categories even if some categories may afterward have no terms, i.e., if all their terms conflicted with other categories.

Second, to resolve conflicting terms, we estimate the quality of all ``clean'' categories.  We calculate a quality score $Q$ for each category (see Section~\ref{sec:anea}; if $|EC| = 0$, then $Q=0$) and sort categories by decreasing $Q$. Sorting brings forward categories that are the most probable to become final categories.

Third, we resolve all conflicting terms by beginning with those that belong to the categories with the highest $Q$. For each conflicting term $t_j$ and all $EC_i$ from which the term originated, we calculate a similarity score $S$: 
$S(t_j, EC_i) = T + L$, where $T$ is the mean cosine similarity between the vector representation of $t_j$ and the remaining terms in a ``clean'' $EC_i$; $L$ is the cosine similarity between $t_j$ and a label of $EC_i$. Even if a ``clean'' $EC_i$ contains no terms, i.e., $T=0$, $L$ will always yield $S>0$. We select the best category for a given term $t_j$ as:
\[EC_{best}(t_j) = \{EC_i| \exists i \in A : \operatorname*{arg\,max} S_i(t_j)\}\]

We add the resolved terms to their best matching category. The final categories are where $|EC| \geq 5$
% \footnote{Figure~\ref{img:anea}d depicts the smaller allowed $|EC|$ due to a very small number of initial terms in the example.}
, i.e., that represent a sufficiently large number of extracted terms from the given domain-specific texts.

%F: a summarizing sentence of this whole section / approach would help to set in context what was the beinning and what is the output and to bringg the reader from the low-level technical description to a higher level of the overall approach, before the readers continue reading the eval

\section{Experiments}
\label{sec:eval}

The evaluation goals are twofold. First, we seek to quantitatively assess the quality of the automatically extracted and annotated terms of both ANEA and a baseline using ratings from domain experts. Second, we seek to identify a recommendation for ANEA's default configuration to automatically annotate texts also of other domains by evaluating the annotated and human-assessed annotated datasets against silver quality datasets.  

Due to the lack of German datasets for the analysis of domain-specific NER, we assess the quality of the produced categories through user studies where we ask users to rate the quality of the entities extracted by our system. We test ANEA and compare it to a baseline on four text datasets with four configurations. 
% TODO evaluation of the domain upsupervised method on the domain annotated english corpus \cite{\cite{zhang2013unsupervised}}

%F: caption is incomplete currently

\subsection{User study}
Our user study aimed at human assessment of the semantic quality of the produced categories due to different configurations. We collect feedback from human assessors for multiple configurations of two methods: ANEA and hierarchical clustering (see Section \ref{sec:hc}). First, we use this feedback to automatically construct silver-quality datasets and evaluate the proposed input configurations against them. Second, we use these silver datasets and evaluate the obtained configurations of the dataset for ANEA to find parameters for default configuration with which ANEA could be used to annotate other domain datasets.    

% if space permits (eg camera ready variant) we should point out the limitations of this type of evaluation, e.g., we effectively measure only "precision" if i understand it correctly? but if a term is not extracted in the first place, it wont be evaluated. 

\subsubsection{Test datasets}
We create four text datasets of comparable size from three different domains: processing industry (P), computer science, and traveling (T). To enable both cross- and intra-domain evaluation, we create two text datasets related to the computer science domain: databases (D) and software development (S). Table~\ref{tab:events} provides an overview of the datasets' parameters, such as the overall number of words, the number of unique terms and heads of terms (see Sec.~\ref{sec:prepr}), and the number of human assessors per each dataset. The table shows that the number of the unique heads may vary even given the identical number of unique extracted terms (cf. datasets S and T ).

%F: changed it so that all should be right aligned except for the first column. somehow, D and T arent, could not fix it though
\begin{table}[h]
\small
\caption{Dataset statistics: databases (D), software development (S), travelling (T), and processing industry (P). The histograms show the distribution of the user-assigned relatedness score to categories extracted with various input configurations per dataset. Bold scores indicate a threshold used to construct silver datasets. }
\centering
\begin{tabular}{lllll}
\hline
     \textbf{Dataset} &  \textbf{D} & \textbf{S} & \textbf{T} & \textbf{P} \\
     \hline
     All words  & 8161 & 8581 & 6293 & 7984\\
     Terms  & 1209 & 1041 & 1040 & 552\\
     Heads& 713 & 673 & 801  & 328 \\ 
     Assessors & 3 & 3 & 2 & 4 \\
     \hline
    \makecell[l]{User study: \\ 1.  cross-term \\relatedness \\ scores} &  \makecell[l]{\includegraphics[width=0.07\textwidth, height=0.07\textwidth]{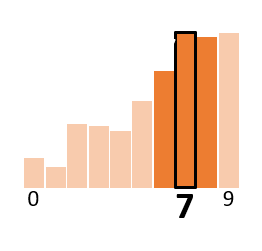}} &  \makecell[l]{\includegraphics[width=0.07\textwidth, height=0.07\textwidth]{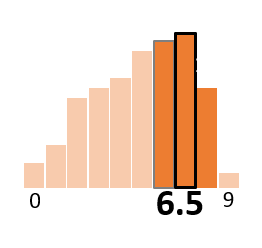}} &  \makecell[l]{\includegraphics[width=0.07\textwidth, height=0.07\textwidth]{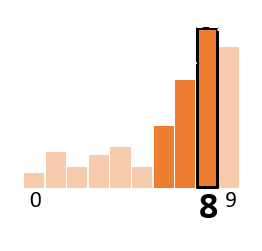}} &  \makecell[l]{\includegraphics[width=0.07\textwidth, height=0.07\textwidth]{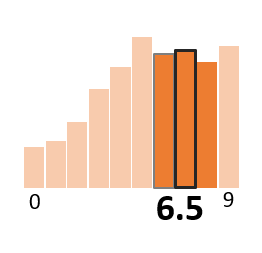}}
    \\
     \makecell[l]{2.   label-terms \\relatedness \\ scores} &  \makecell[l]{\includegraphics[width=0.07\textwidth, height=0.07\textwidth]{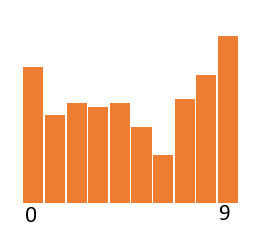}} &  \makecell[l]{\includegraphics[width=0.07\textwidth, height=0.07\textwidth]{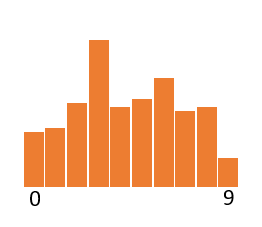}} &  \makecell[l]{\includegraphics[width=0.07\textwidth, height=0.07\textwidth]{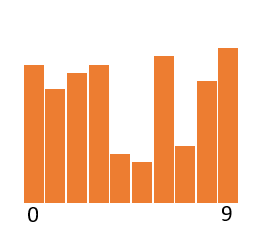}} &  \makecell[l]{\includegraphics[width=0.07\textwidth, height=0.07\textwidth]{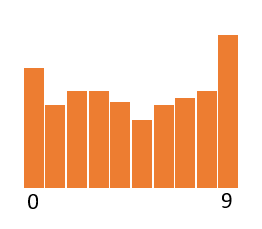}}
     \\
     \hline
\end{tabular}
% \caption{Characteristics of the test datasets. }
\label{tab:events}
\end{table}

To test the applicability of the approach on different domains, we use the publicly available data from Wikipedia and a dataset built on private text data from a real-world production line in the processing industry. Specifically, the first three datasets (databases, software development, and traveling) originate from German Wikipedia articles dedicated to the respective categories. For each dataset, we searched for related articles in Wikipedia using a query ``incategory:\textit{category}'', where ``category'' is ``Datenbanken,'' ``Programmierung,'' or ``Reise''. We iterated over the list of the search results sorted by relevance and extracted the texts of the articles if the articles had a specific number of words $W$: $220 \leq W \leq 2500$, i.e., articles of medium size. The last dataset consists of reports about the daily operations of a company in the processing industry. Such reports include texts about statuses of the machinery, processes in the production lines, and problems that occurred throughout the daily routines. The dataset consists of approximately 200 short texts, each of 20-100 words. 
% We stopped adding texts as soon as the number of words from the new articles would exceed 9000 or 7000. 

\begin{figure*}[h]
\centering
\includegraphics[width=0.95\textwidth]{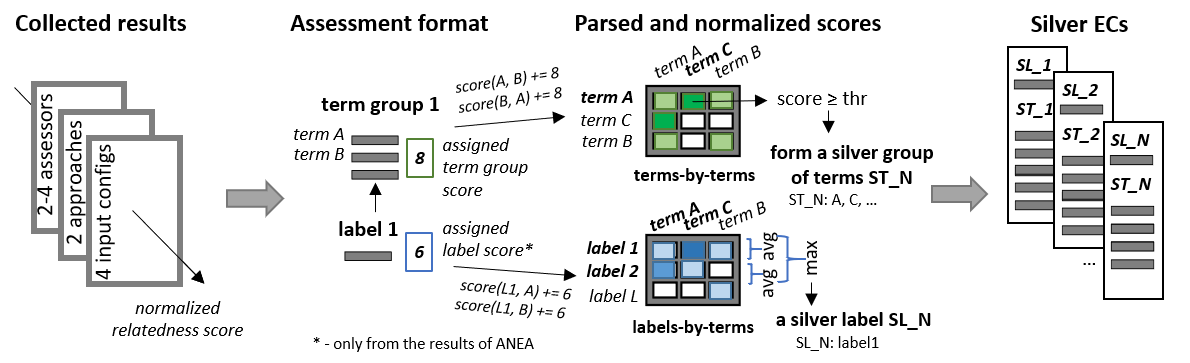}
\caption{\label{img:silver} \centering Construction of a silver dataset through collecting of information from user studies.}
\end{figure*}

\subsubsection{Experiment setup} 
For the human assessment, we recruited nine native-German speaking participants (4 f, 5 m, aged between 23-60). Each participant is familiar with the domain of the assigned dataset(s) through their job, education, and/or hobbies. 

We assigned 3-4 participants to each dataset, and each participant evaluated one or two datasets. Albeit the processing industry dataset has the smallest number of unique terms, we assigned the largest number of assessors to it due to the high relevance of obtaining valid results for such complex, expert domains as chemistry and technology. The vocabulary of these domains is typically strongly underrepresented in general text corpora used to train word embedding models \cite{grave2018learning}. 

The evaluation included two tasks for the participants:  (1) assess the cross-term relatedness within the identified groups of terms and (2) assess the relatedness of the labels automatically assigned to the identified groups of terms. Per dataset, each participant needed to perform an assessment of eight sheets with automated annotation results: four identical input configurations per both ANEA and a baseline. Each participant needed to assign a semantic relatedness score between 0 and 9, where 0 meant no similarity and 9 - the highest similarity.

The input configuration included four different numbers of the input terms, i.e., \textit{terms-to-annotate (TTAs)}, among which the algorithms needed to extract the most representative terms that can form a separate semantic concept, i.e., an category. To vary the size of TTAs, we selected $1/Z \cdot 100\%$ most frequent heads of phrases:  $Z\in[2,3,4,5]$ for the datasets with a number of unique terms $< 1000$, else $Z\in[3,4,5,7]$. By selecting only terms that share all most frequent heads, we ensure that these terms are the most representative of each domain-specific text. 

Table \ref{tab:events} reports the results of the user studies and shows that the relatedness of the groups of terms is higher than assigned labels. That is, the cross-term relatedness score was biased towards higher values, the label relatedness had a more uniform score distribution. Additionally, the mean and maximum of the relatedness scores vary across the datasets. We noticed that the relatedness scores were biased toward the size of the identified categories, i.e., a smaller number of terms in categories tend to have higher scores since it is easier for a human to assess a smaller number of items. However, we did not find any correlation between individual datasets and any of the outlined numeric characteristics. 

To estimate which input configuration and approach yielded the most coherent categories, we require a silver dataset, which will average the assigned scores and extract the highly rated combinations of terms into categories, and assigned the highly rated labels to them.

\subsubsection{``Silver'' datasets} The goal of a silver dataset is to ensure a fair and unified evaluation strategy of the approaches for all topics. We constructed a silver dataset for each topic by aggregating information from the human assessment sheets following the identical procedure. 

First, for each dataset, we constructed term-to-term and label-to-terms score matrices between the vocabulary of each topic and the extracted and assigned labels (Figure \ref{img:silver}). The matrices were initialized with zeros. We iterated over the relatedness scores across two approaches, four input configurations, and two-four human assessors. For every two terms in a term group, we added an assigned cross-term relatedness score to a value in a term-to-term matrix. This score demonstrates how two terms are evaluated in various combinations with other terms across different setups. After the summation was completed, we normalized each value in the matrix by the number of times two terms occurred together. We performed a similar procedure with the label-to-term relatedness scores: for each term in a category and an assigned label to a category, we added a label-to-terms relatedness score assigned to a category and then normalized by the number of times a label was applied to a term. 

To identify a threshold of relatedness of two terms belonging to a category, we built a histogram of all scores used to evaluate each dataset (see Table~\ref{tab:events}). In a score range of 0-9, we decided that a threshold of sufficient relevance of terms needs to lie higher than the mean score and not equal to the maximum value, i.e., between scores 6-8. Thus, for each dataset, we chose the most frequent score as a threshold and if a preceding score was less frequent by 1, then we calculated a mean of these scores. 

We collected silver groups of terms, by choosing a term and merging it with other terms, a normalized relatedness score to which is higher or equal to the threshold of this dataset. We expanded this list of terms with the terms that are related to any of the merged terms compared to the threshold. Note that relatedness of all terms to each other exceeded a relatedness threshold, but relatedness of at least two terms needed to exceed a threshold. If a group of terms contained at least five terms, we form a silver category. We assigned a label to a silver category by (1) calculating mean label-to-terms scores of all labels applied to at least two identified terms of a group, (2) selecting a label with the maximum mean score.

\subsection{Evaluation}
To evaluate the coherence and semantic quality of the produced categories at various input configurations, we introduce the evaluation methodology to evaluate ANEA and a representative baseline against the silver datasets. By identification the input configurations that yielded the best results, we sought to propose an optimal default ANEA's input configuration for any dataset.

\subsubsection{Baseline: hierarchical clustering}
\label{sec:hc} We selected hierarchical clustering (HC) \cite{murtagh2012algorithms} as a baseline to ANEA, since it successfully identifies semantically related terms that refer to identical entities \cite{cambria2016senticnet}. Although HC does not have the functionality of automated extraction and assignment of a label of a cluster of terms, we observed that HC's clusters could form meaningful categories. Therefore, we selected HC as a baseline to compare the quality of the produced groups of terms. 

To ensure the best performance of HC per each dataset and each input configuration, we created an optimization of HC that selects the best clustering results and outputs clusters that contain maximum terms with maximum cross-term similarity. For each group of terms, we ran HC four times with fixed parameters of cosine similarity and average linkage criterion. We chose the linkage method, distance metric, and the optimization of the hyperparameters for the HC that are the most similar to the ANEA. 

We built clustering configurations by varying the similarity threshold value between $[0.5; 0.8]$ with a step of 0.1. For each clustering configuration $j$, we selected only clusters $CL_{j,i}$ with more than 5 terms in each ($|CL_{j,i}| \geq 5$), i.e., impose the same minimum size requirements as for ANEA. Then, we calculated a weighted similarity score of each parameter configuration $WS_j$: 
\[WS_j = \frac{1}{\sum_{i=0}^I |CL_{j,i}|} \sum_{i=0}^I T_{j,i} \cdot |CL_{j,i}|\]
where $I$ is the number of clusters larger than 5 produced at a run $j$, and $T_{j,i}$ is a cross-term similarity within a cluster. We selected the best configuration as 
$C_{best} = \operatorname*{arg\,max}_j (WS_j \cdot \sum_{i=0}^I |CL_{j,i}|)$, i.e., a configuration that clusters the most terms and in the most semantically coherent way.

Since HC does not have label extraction and assigning functionality, the human assessors received only one task of assessment of the cross-term relatedness of clusters produced by HC. 

\subsubsection{Metrics} To evaluate the quality of the identified categories, we use five parameters: (1) number of categories: a larger number indicates diverse and narrowly defined categories, a smaller number - generalizing categories, (2) number of annotated terms (AT): property of identified relations between more terms, (3) the average size of categories: a smaller size indicates more narrowly-defined categories, whereas a larger size indicates more generally-related terms in categories; (4) average cross-term score (TS), (5) average label-to-terms score (LS), and (6) average score (AS) between TS and LS: the high scores indicate the higher relatedness of the extracted terms and extracted and assigned labels. The main goal of our evaluation is to identify which input configurations lead to the highest average score between cross-term and label-to-terms relatedness while annotating more TTA into more general categories.

\begin{table}[]
\caption{Evaluation of the entity identification methods in various input configuration. Z is the denominator to choose terms-to-annotate from the most $1/Z$ frequent terms' heads; TTA is the number of \textit{terms-to-annotate}, EC is the number of the \textit{(entity) categories} that were produced by each approach; AT is the number of \textit{annotated terms}, i.e., that belong to the identified categories; TS is the mean cross-\textit{term relatedness score} among the categories; LS is the mean \textit{label-to-terms relatedness score}; AS is the \textit{average score} between TS and LS (* means that AS is equal to TS because HC does not assign labels to groups of terms).}
\begin{tabular}{l|l|l|l|l|l|l|l|l|l}
topic              & Appr.                  & Z & TTA & EC & AT & Size & TS  & LS  & AS\\
\hline
\multirow{9}{*}{D} & silver                & 3     & 420 & 5   & 113 & \textbf{23} & 7.2 & 7.0 & \textbf{7.2}\\
                    \cline{2-10}
                    % & freq.heads                & 3     & 420 & 13   & 88 & 6.6 & 3.7 & 5.1\\
                    % \cline{2-9}
                   & \multirow{4}{*}{HC}   & 3     & 420 & 14  & 108 & 8 & 6.9 & -- & 6.9* \\
                   &                       & 4     & 363 & 12  & 87  & 7 & 7.0 & -- & 7.0*\\
                   &                       & 5     & 316 & 10  & 73  & 7 & 7.0 & -- & 7.0*\\
                   &                       & 7     & 253 & 8   & 52  & 7 & 7.2 & -- & \textbf{7.2}* \\
                   \cline{2-10}
                   & \multirow{4}{*}{ANEA} & 3     & 420 & 26  & \textbf{306} & 12 & 4.7 & 4.2 & 4.4 \\
                   &                       & 4     & 363 & 22  & 255 & 12 & 5.3 & 4.6 & 5.0 \\
                   &                       & 5     & 316 & 21  & 234 & 11 &  5.6 & 5.0 & 5.3\\
                   &                       & 7     & 253 & 18  & 179 & 10 & 5.7 & 5.0 & 5.4\\
\hline
\multirow{9}{*}{S} & silver                & 3     & 356 & 6   & 57  & 10 & 6.2 & 6.0 & \textbf{6.1}\\
                   \cline{2-10}
                %   & freq.heads             & 3     & 356 & 4   & 24 & 5.6  & 5.8 & 5.7\\
                %     \cline{2-9}
                   & \multirow{4}{*}{HC}   & 3     & 356 & 17  & 190 & 11 & 5.3 & -- & 5.3*\\
                   &                       & 4     & 303 & 15  & 152 & 10 &  5.5 & -- & 5.5*\\
                   &                       & 5     & 255 & 15  & 137 & 9 & 5.4 & -- & 5.4*\\
                   &                       & 7     & 191 & 12  & 103 & 9 & 5.4 & -- & 5.4*\\
                   \cline{2-10}
                   & \multirow{4}{*}{ANEA} & 3     & 356 & 21  & \textbf{242} & 12 & 4.1 & 4.4 & 4.2\\
                   &                       & 4     & 303 & 18  & 186 & 10 & 4.2 & 3.8 & 4.0\\
                   &                       & 5     & 255 & 15  & 164 & 11 & 4.2 & 4.4 & 4.3\\
                   &                       & 7     & 191 & 10  & 119 & 12 & 5.0 & 5.3 & 5.2\\
\hline
\multirow{9}{*}{T} & silver                & 3     & 363 & 6   & 115 & 19 & 7.8 & 6.7 & \textbf{7.3}\\
                   \cline{2-10}
                %   & freq.heads                & 3     & 363 & 2   & 10 & 7.7 & 8.0 & 7.8\\
                %     \cline{2-9}
                   & \multirow{4}{*}{HC}   & 3     & 363 & 19  & 156 & 8 & 7.3 & -- & \textbf{7.3}*\\
                   &                       & 4     & 297 & 16  & 133 & 8 & 7.3 & -- & \textbf{7.3}*\\
                   &                       & 5     & 258 & 12  & 105 & 9 & 7.3 & -- & \textbf{7.3}*\\
                   &                       & 7     & 211 & 9   & 76  & 8 & 7.2 & -- & 7.2*\\
                   \cline{2-10}
                   & \multirow{4}{*}{ANEA} & 3     & 363 & 22  & \textbf{239} & 11 & 5.4 & 4.8 & 5.1\\
                   &                       & 4     & 297 & 17  & 191 & 11 & 5.4 & 4.4 & 4.9 \\
                   &                       & 5     & 258 & 14  & 161 & 12 & 5.3 & 4.4 & 4.9 \\
                   &                       & 7     & 211 & 14  & 137 & 10 & 5.6 & 4.0 & 4.8 \\
\hline
\multirow{9}{*}{P} & silver                & 2     & 282 & 7   & 102 & 15 & 6.6 & 6.2 & \textbf{6.4}\\
                   \cline{2-10}
                %   & freq.heads                & 2     & 282 & 7   & 58 & 6.4 & 2.5 & 4.5\\
                %     \cline{2-9}
                   & \multirow{4}{*}{HC}   & 2     & 282 & 9   & 65 & 7  & 5.2 & -- & 5.2* \\
                   &                       & 3     & 227 & 8   & 61 & 8 & 6.0 & -- & 6.0*\\
                   &                       & 4     & 200 & 8   & 61 & 8 & 6.0 & -- & 6.0*\\
                   &                       & 5     & 183 & 7   & 56  & 8 & 6.1 & --  & 6.1*\\
                   \cline{2-10}
                   & \multirow{4}{*}{ANEA} & 2     & 282 & 18  & \textbf{213} & 12 & 4.7 & 4.5 & 4.6 \\
                   &                       & 3     & 227 & 16  & 172 & 11 & 5.3 & 4.9 & 5.1\\
                   &                       & 4     & 200 & 16  & 163 & 10 & 5.3 & 4.8 & 5.1\\
                   &                       & 5     & 183 & 15  & 149 & 10 & 5.2 & 4.5 & 4.8 \\

\end{tabular}
\label{table:eval1}

\end{table}

\subsubsection{Results}
To calculate average relatedness scores, we assigned the scores from the normalized score matrices to the terms and labels of the identified EC and average these scores.   Table \ref{table:eval1} reports the evaluation results for four datasets and four input configurations. The table shows that HC gets the highest average relatedness score ($AS_{avg, HC} = 6.5$) that almost reaches silver dataset ($AS_{avg,silver} = 6.7$) but at the same time produces categories of the size smaller than silver categories ($Size_{avg, HC} = 6$ and $Size_{avg, silver} = 16$). While on average, ANEA annotates the largest number of terms ($AT_{mean, ANEA} = 175$), it also yields the lowest average relatedness score ($AS_{mean, ANEA} = 5.2$) both compared to the silver dataset and HC. When creating a coding book, multiple human coders first annotate, and then the majority voting decides which excepts and labels describe a dataset the best. We applied a similar strategy to improve the performance of ANEA.

\begin{table}[]
\caption{Evaluation of the entity identification methods when (entity) categories (EC) are created by majority voting between input configurations. In each dataset, the majority voting improves relatedness score for each dataset. The highlighted configurations show a range of terms-to-annotate (TTA) and the resulted best average score (AS). }

\begin{tabular}{l|l|l|l|l|l|l|l|l}
              & Conf. Z       & TTA      & EC & AT & Size & TS   & LS   & AS  \\
\hline
\multirow{8}{*}{D}  & silver& 420 & 5   & 113 &23 & 7.2 & 7.0 & 7.2\\
                    & prev best: 7     & 253 & 18  & 179 & 10 & 5.7 & 5.0 & 5.4\\
                    \cline{2-9}
                    & 3+4     & 363-420 & 16  & 160 & 10 & 5.7 & 5.1 & 5.4 \\
                   & 3+4+5   & 316-420 & 16  & 230 & 14 & 5.5 & 4.9 & 5.2 \\
                   & 3+4+5+7 & 253-316 & 16  & 247 & 15 & 5.4 & 4.8 & 5.1 \\
                   & 4+5     & 316-363 & 16  & 172 & 11 & 6.0 & 5.5 & 5.7 \\
                   & 4+5+7   & 253-363 & 15  & 202 & 13 & 5.7 & 5.1 & 5.4 \\
                   & 5+7     & \underline{\textbf{253-316}} & 12  & 122 & 10 & 6.3 & 5.9 & \underline{\textbf{6.1}} \\
\hline
\multirow{8}{*}{S} & silver & 356 & 6   & 57  & 10 & 6.2 & 6.0 & 6.1\\
                & prev best: 7     & 191 & 10  & 119 & 12 & 5.0 & 5.3 & 5.2\\
                \cline{2-9}
                    & 3+4     & 303-356 & 10  & 95  & 10 & 4.5 & 5.0 & 4.8 \\
                   & 3+4+5   & 255-356 & 13  & 173 & 13 & 4.4 & 4.6 & 4.5 \\
                   & 3+4+5+7 & 191-356 & 11  & 185 & 17 & 4.4 & 4.4 & 4.4 \\
                   & 4+5     & 255-303 & 8   & 89 & 11 & 4.4 & 4.5 & 4.4 \\
                   & 4+5+7   & 191-303 & 10  & 126 & 13 & 4.2 & 4.8 & 4.5 \\
                   & 5+7     & \underline{\textbf{191-255}} & 4   & 44  & 11 & 5.6 & 6.5 & \underline{\textbf{6.0}} \\
\hline
\multirow{8}{*}{T}  & silver   & 363 & 6   & 115 & 19 & 7.8 & 6.7 & 7.3\\
                & prev best: 3     & 363 & 22  & 239 & 11 & 5.4 & 4.8 & 5.1\\
                \cline{2-9}
                    & 3+4     & 297-363 & 14  & 119 & 9 &  6.1 & 5.3 & 5.7 \\
                   & 3+4+5   & \underline{\textbf{258-363}} & 12  & 146 & 12 & 6.2 & 5.6 & \underline{\textbf{5.9}} \\
                   & 3+4+5+7 & 211-363 & 9   & 171 & 19 & 6.1 & 5.6 & 5.8 \\
                   & 4+5     & 258-297 & 10  & 101 & 10 & 6.1 & 5.0 & 5.6 \\
                   & 4+5+7   & 211-297 & 12  & 137 & 11 & 5.8 & 5.1 & 5.4 \\
                   & 5+7     & 211-258 & 9   & 86 & 10 & 5.9 & 5.2 & 5.6 \\
\hline
\multirow{8}{*}{P} & silver &  282 & 7   & 102 & 15 & 6.6 & 6.2 & 6.4\\
                    & prev best: 3     & 227 & 16  & 172 & 11 & 5.3 & 4.9 & 5.1\\
                     \cline{2-9}
                    & 2+3     & 227-282 & 12  & 133 & 11 & 5.6 & 5.6 & 5.6 \\
                   & 2+3+4   & 200-282 & 14  & 160 & 11 & 5.5 & 5.2 & 5.4 \\
                   & 2+3+4+5 & \underline{\textbf{181-282}} & 9   & 157 & 17 & 5.7 & 5.6 & \underline{\textbf{5.6}} \\
                   & 3+4     & 200-227 & 12  & 128 & 11 & 5.5 & 4.9 & 5.2 \\
                   & 3+4+5   & 183-227 & 14  & 146 & 10 &  5.4 & 4.7 & 5.0 \\
                   & 4+5     & 200-227 & 13  & 121 & 9 & 5.4 & 5.0 & 5.2 \\

\end{tabular}
\label{table:eval2}
\end{table}

\subsubsection{Voting strategy and default input configuration}
Ensemble learning is a common approach in machine learning to improve the results of a classifier, i.e., by combining predictions of multiple classifiers to achieve a boost in the overall accuracy through collecting ``the wisdom of the crowd'' \cite{berthold2010guide}. 

We followed this principle to improve the quality of the extracted by ANEA categories and combine results of 2-4 configurations in each dataset. Similar to the construction of silver datasets, we created a category of at least five terms if these terms co-occurred in at least two input configurations. We assigned a label that describes the majority of the terms in the identified group. 

Table \ref{table:eval2} reports that at least one of the combination of ANEA with multiple input configurations increases the average relatedness score compared to ANEA without a voting strategy on average by 0.7 ($AS_{avg, ANEA_{vote}} = 5.9$). Although the voting approach does not exceed the relatedness scores of the silver datasets (reaches 87.9\% of the silver score), it increases a number of annotated terms per category ($AS_{avg, ANEA} = 11$ and $AS_{avg, ANEA_{vote}} = 13$) and also identifies more generalizing categories ($EC_{avg, ANEA} = 15$ and $EC_{avg, ANEA_{vote}} = 9$).

To identify the best default configuration for the voting strategy of ANEA, we selected the best performing voting strategy configurations per each dataset and deduced a default input configuration by generalizing these configurations. We took the minimum and the maximum number of terms-to-annotate (TTAs) from the best voting configuration and plot against a number of unique heads in each dataset (see Table \ref{tab:events}). 

\begin{figure}[h]
\centering
\includegraphics[width=0.45\textwidth]{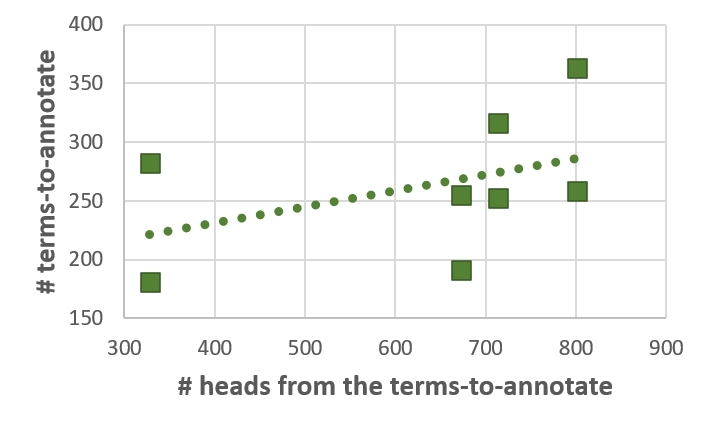}
\caption{\label{img:trend} \centering Default input configuration: a number of terms-to-annotate linearly depends on the number of unique NP heads per dataset. }
\end{figure}

Figure \ref{img:trend} depicts a linear trend between the TTAs and the unique heads from the datasets. Based on this trend, for any other dataset, we recommend annotating only the first $y=158 + 0.167x$ TTAs that belong to the most frequent heads, where $x$ is a number of unique heads in a dataset. For the voting strategy, we recommend using the input configurations of the $y$, $y-40$, and $y+40$ number of terms that share the most frequent heads of terms.

\section{Discussion and future work}

Our evaluation shows that ANEA facilitates a faster annotation process. Specifically, ANEA automatically performs the most time-consuming tasks of deriving a coding book for the annotation of a dataset for NER. ANEA imitates the first two stages of a manual annotation process. First, a small set of articles ($\approx$6000-8000 words) is used to automatically identify categories relevant to the data of the current domain, i.e., identify and extract related terms and assign a label to each of them (identified 4-12 categories with 44-157 assigned terms). Second, a voting strategy is applied, which aims at increasing the validity of the derived categories by following the idea of ensemble learning and intercoder agreement (relatedness score improved on average from 77.4\% of the silver average relatedness score to 87.9\%). To continue with manual annotation of a (N)ER dataset, next, researchers manually validate their coding book. If they find that the coding book sufficiently represents the dataset, they annotate the remaining texts to create a large corpus for NER. 

Therefore, the primary use cases of ANEA are as follows. First, extraction of domain categories from a subset of a large text dataset and improve their quality with the voting strategy. Second, manual validation and improvement of the identified categories by moving terms between the categories and suggesting better labels to them.  

The final stage of annotation of a NER dataset is to apply a coding book to a large dataset, i.e., read the text and assign categories to text excerpts following guidelines or examples of the coding book. Although, such manual text annotation is a standard approach to create ``gold''-standard datasets, the recent semi-supervised learning neural network models (e.g., DART \cite{chang-etal-2020-dart}) show high potential in reliable annotating of large text collections. We plan to use the models like DART will complete the automation of creating domain-specific NER datasets.

Future work directions include the creation of manually annotated datasets from scratch for multiple domains to calculate accuracy metrics, e.g., precision, recall, and F1, to evaluate the effectiveness of identifying categories by ANEA. Further, to improve the quality and meaningfulness of the assigned labels, we plan to test ANEA on other knowledge graphs, e.g., Wikidata or BabelNet. To test the applicability of ANEA, we also plan to evaluate the approach in other languages, e.g., English, with an additional module for the identification of multi-word expressions similar to compound-based German words \cite{salehi2014detecting}. Further, to improve the semantic quality of both categories' terms and labels in a specific domain, we plan to use a language model, e.g., BERT, use the quality score as a learning objective. Lastly, we seek to build a semi-supervised NER model to complete automated annotation of NER datasets, i.e., automatically annotate large datasets suitable for training neural network models. We will use terms and labels from the derived categories as the seed-terms and seed-labels and perform named entity tagging and classify more domain terms \cite{shang2018learning, chang-etal-2020-dart}.

\section{Conclusion}
In this paper, we propose ANEA, an automatic approach to derive domain entity categories from a subset of domain input texts, i.e., create a small dataset to train a NER model. Specifically, ANEA identifies related domain-representative terms and automatically extracts and assigns descriptive and generalizing labels to them based on Wiktionary. In our user assessment and evaluation, ANEA could not outperform a silver dataset on the relatedness scores assigned to the groups of terms and labels describing these groups. However, ANEA produced more generalizing domain categories compared to a strong baseline. We showed that our voting strategy of combining terms and labels from the categories identified at multiple input configurations significantly improved the quality of the final categories. Additionally, we suggested a default input configuration that can be applied to derive categories from German domain text datasets. Finally, we think that the best application of ANEA is to annotate and use a small dataset in semi-supervised learning. Moreover, we plan to improve and validate the annotations with a domain expert, and use this small domain dataset to train state-of-the-art NER models.

\section*{Acknowledgement}
The research for this paper has been conducted in collaboration with the company eschbach (\url{https://eschbach.com}) supported by the Central Innovation Programme (ZIM) of the German Federal Ministry for Economic Affairs and Energy. 

We thank all study participants for their significant contribution to this publication.

\bibliographystyle{ACM-Reference-Format}
\bibliography{sample-base}

\end{document}